%% file: VPR4LQQ.tex
\documentclass[letterpaper, 10 pt, conference]{ieeeconf}   

\IEEEoverridecommandlockouts                   

\makeatletter
\let\NAT@parse\undefined
\makeatother

\usepackage[table,dvipsnames]{xcolor}
\usepackage{tabularx}
\usepackage{booktabs}
\usepackage{stfloats}
\usepackage{tikz}
\usepackage{graphics} 
\usepackage{graphicx}
\usepackage{grffile} 
\usepackage{amsmath} 
\usepackage{amssymb}  
\usepackage{cite}
\usepackage{bm}
\usepackage{multicol}
\usepackage{url}

\usepackage[noend]{algpseudocode}
\usepackage{algorithm}
\usepackage{makecell}
\usepackage{subfig}
\usepackage[square, comma, numbers,sort]{natbib}
\usepackage{multirow}
\let\labelindent\relax
\usepackage{enumitem}
\usepackage{caption}
\usepackage{pifont}
\usepackage{scalefnt}
\usepackage{mwe}
\usepackage{array}
\usepackage[pagebackref=true,breaklinks=true,colorlinks,bookmarks=false]{hyperref}
\hypersetup{
linkcolor=BrickRed
,citecolor=Green
,filecolor=Mulberry
,urlcolor=NavyBlue
,menucolor=BrickRed
,runcolor=Mulberry
,linkbordercolor=BrickRed
,citebordercolor=Green
,filebordercolor=Mulberry
,urlbordercolor=NavyBlue
,menubordercolor=BrickRed
,runbordercolor=Mulberry
}
\usepackage{balance}

\usepackage{blindtext} % REMOVE: only for testing

\input{Parts/0a-Title_Authors}

\begin{document}

\maketitle
\thispagestyle{empty}
\pagestyle{empty}

\input{Parts/0b-Abstract}
\input{Parts/1-Introduction}

\input{Parts/2-RelatedWork}
\input{Parts/3-Method}
\input{Parts/4-Experiment}
\input{Parts/5-Results}
\input{Parts/6-AdditionalExperiments}

\input{Parts/7-Conclusion}

\scalefont{0.81}
\bibliographystyle{IEEEtran}
\balance
\bibliography{VPR4LQQabrv}

\end{document}

%% file: Parts/0a-Title_Authors.tex
\title{\LARGE \bf
Distillation Improves Visual Place Recognition for Low Quality Images
}

\author{Anbang Yang\textsuperscript{*}, Ge Jin\textsuperscript{*}, Junjie Huang, Yao Wang, John-Ross Rizzo, Chen Feng\textsuperscript{\ding{41}}\vspace{-6mm}\\% <-this % stops a space
\thanks{\textsuperscript{*} Equal contribution.}
\thanks{New York University, Brooklyn, NY 11201, USA}%
\thanks{\ding{41} Corresponding author (\href{mailto:cfeng@nyu.edu}{cfeng@nyu.edu}). This work is supported partly by NSF Grants 2238968 and 2345139; by the National Eye Institute and Fogarty International Center under Grants R21EY033689, R33EY033689, and R01EY036667; and by the NYU IT High Performance Computing resources, services, and staff expertise.}%
}

%% file: Parts/0b-Abstract.tex
\begin{abstract}

Real-world applications of Visual Place Recognition (VPR) often rely on cloud computing, where query images or videos are transmitted elsewhere for visual localization. However, limited network bandwidth forces a reduction in image quality, which degrades global image descriptors and consequently VPR accuracy. We address this issue at the descriptor extraction level with a knowledge-distillation framework that transfers feature representations from high-quality images to low-quality ones, allowing VPR methods to produce more discriminative descriptors from the latter. Our approach leverages three complementary loss functions ---- the Inter-channel Correlation Knowledge Distillation (ICKD) loss, Mean Squared Error (MSE) loss, and a weakly supervised Triplet loss ----- to guide a student network in approximating the high-quality features produced by a teacher network. Extensive experiments on multiple VPR methods and datasets subjected to JPEG compression, resolution reduction, and video quantization demonstrate significant improvements in recall rates. Furthermore, our work fills a gap in the literature by exploring the impact of video-based data on VPR performance, thereby contributing to more reliable place recognition in resource-constrained environments. Source code and data are available at \href{https://ai4ce.github.io/LoQI-VPR/}{https://ai4ce.github.io/LoQI-VPR/}.

\end{abstract}

%% file: Parts/1-Introduction.tex
\section{Introduction}\label{introduction}

Visual Place Recognition (VPR) has rapidly become a foundational component in machine perception. Through learning compact vector representations of RGB images, or global image descriptors, VPR methods localize unseen images by recalling one or more known locations among a database of images based on descriptor similarity \cite{garg2021your,masone2021survey}. VPR is critical for real-world applications ranging from robotic navigation in complex industrial environments to augmented reality systems and way finding aids for vulnerable populations \cite{doan2019scalable,yin2019multi,xin2019localizing}. 

However, VPR deployment at scale comes with an inherent dilemma between VPR performance and computational / networking costs. On one hand, state-of-the-art (SOTA) VPR methods leverage deep learning to produce highly discriminative and generalizable global descriptors \cite{ali2023mixvpr,lu2024cricavpr}. However, such methods are computationally costly for both descriptor extraction and matching \cite{garg2020fast}, making large-scale deployment on consumer / edge devices impractical. Besides computational cost, there is also a further dilemma caused by the storage demands of the VPR image database, which is difficult to maintain locally given the rapid growth in scene diversity and volume for real-world VPR applications \cite{kornilova2023dominating,berton2022rethinking}.

\begin{figure}[t]
	\centering
	\includegraphics[width=1\columnwidth]{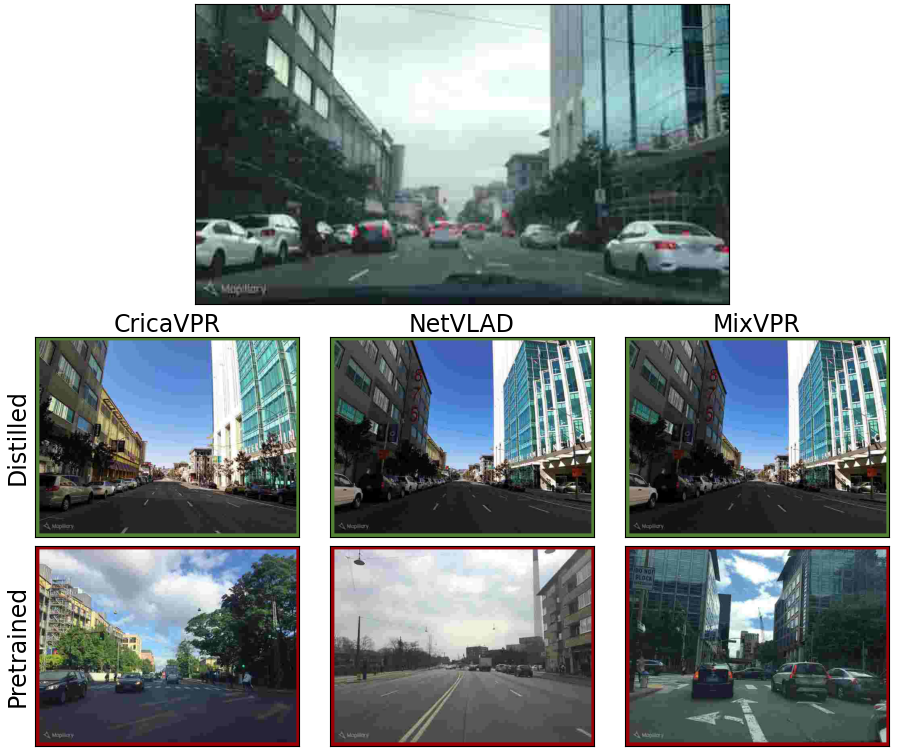}
    \caption{\textbf{Enhanced Low Quality Image Retrieval:} After knowledge distillation from high quality images, this JPEG-compressed query image from the Mapillary SLS dataset (top center) is correctly localized by multiple VPR methods (top row), whereas they failed beforehand (bottom row).}
	\label{fig:overview}
	\vspace{-15pt}
\end{figure}

Given these challenges, one option is to perform VPR locally on low-end devices using methods specifically designed for low computational cost. The second option is to perform VPR over the cloud, where images are transmitted elsewhere for centralized processing. Arguably, the cloud VPR approach has greater potential for accurate localizations, as it carries the aforementioned benefits of SOTA VPR methods. However, network bandwidth constraints are common when many devices simultaneously capture images to query the VPR database, hampering real-time communication with servers. As most real-world VPR applications localize over image sequences, the underlying video streaming service could alleviate bandwidth pressures by reducing resolution and increasing image or video compression. However, such quality reductions to VPR methods' visual inputs significantly hamper localization accuracy \cite{tomita2023visual}.

Our work offers a potential solution to the dilemmas of real-world VPR, benefiting from the performance and generalizability of SOTA VPR methods while mitigating VPR accuracy loss from lowered image quality. As networking-induced image quality reduction within the cloud VPR paradigm is not easily eliminated, we propose a knowledge distillation framework to improve the performance of VPR methods under low-quality images. Although image processing techniques not specific to VPR such as super-resolution \cite{park2023content,yue2023resshift} and image deblurring \cite{kupyn2018deblurgan,kong2023efficient} can potentially restore images' visual contents after network transmission, these methods do not account for images' role as place representations and thus are not directly beneficial for VPR performance. In short, we offer the following contributions:
\begin{itemize}
    \item Utilizing knowledge distillation, we develop a training framework to improve the localization performance of VPR global descriptors extracted from low-quality images. Our method is not limited to any particular VPR method.
    \item We validate our framework across multiple public VPR datasets and various types of low-quality images, including JPEG compression, resolution reduction, and video quantization. In doing so, we also curate a video-based VPR dataset to address the low availability of video data for VPR research.
\end{itemize}

% {\color{gray}
% Outline of the introduction
% \begin{itemize}
%     \item talk about modern VPR methods \cite{ali2023mixvpr, lu2024cricavpr} etc.
%     \item SOTA methods are computationally intensive \cite{garg2020fast}, and consumer devices have little storage / compute
%     \item two choices: 1. run locally, worse performance 2. run remotely, slower response due to networking bandwidth limit
% \end{itemize}

% We would like to bypass this dilemma altogether.
% \begin{itemize}
%     \item Yes we will use SOTA methods, but...
%     \item traditionally people use choice \#2 but with lower quality images, which is bad for VPR performance
%     \item We also use choice \#2; however, we would like to improve performance even given lower quality images
% \end{itemize}

% \textbf{...and that is the point of our paper}

% \textit{but before we talk about what we gonna do, let's first get some naive solutions out of the way.}
% \begin{itemize}
%     \item Super resolution doesn't work (we have experimental evidence to support this as well, a figure from ICRA 2024).
%     \item De-blurring also doesn't work, just not improving performance whatsoever
% \end{itemize}
% }

%% file: Parts/2-RelatedWork.tex
\section{Related Work}

\subsection{VPR Under Low-Quality Image Conditions}\label{low_quality_ways}

% 1. image quality reduction: 2 ways

Prevailing VPR literature frames the problem of image-based localization as an image retrieval task, and research in VPR methods has traditionally focused on improving retrieval accuracy under high-level visual changes such as illumination changes, seasonal variation, and viewpoint shifts \cite{masone2021survey}. However, the impact of \emph{image quality degradation} (e.g., low resolution, heavy compression, or motion blur) has received comparatively less attention. Recent studies have begun to highlight this gap. For instance, the accuracy of deep learning-based VPR methods such as NetVLAD \cite{arandjelovic2016netvlad} suffers significantly from low-resolution images \cite{tomita2023visual}. In addition to lowered image resolution, increased JPEG compression has also been shown to negatively impact VPR methods \cite{tomita2023compress}. Given the high compatibility and widespread adoption of the JPEG image format \cite{hudson2018jpeg}, it is prevalent in image transmission when deploying VPR at scale \cite{yang2022unav}. Furthermore, many widely used VPR datasets for studying the aforementioned high-level retrieval challenges also store images in JPEG format. Examples include Tokyo247 \cite{torii2018tokyo247} (illumination changes), Mapillary SLS \cite{warburg2020msls} (large temporal variation), Nordland \cite{sunderhauf2013nordland} (seasonal variation), and GSV-Cities \cite{ali2022gsv} (viewpoint shifts). Although in our work we consider multiple forms of image quality reduction, we place a special emphasis on JPEG compression for its relevance to real-world deployments and existing VPR datasets.

Besides image quality reduction, the cloud VPR paradigm also faces the challenge of video compression. Within VPR research, localization accuracy under video compression and specifically video quantization is not extensively studied. Even lightweight VPR methods \cite{garg2020fast,ferrarini2022floppy,arcanjo2022droso}, the alternative solution to the dilemmas of computational cost / storage demands introduced in \ref{introduction}, still focus on the image domain. As video quantization introduces additional artifacts that further degrade image quality, we seek to better understand video compression's effects on VPR descriptor extraction. To our best knowledge, we also contribute towards a literature gap on publicly available datasets offering video data.

% 2. video compression

% To mitigate performance loss under low-quality image conditions, a common approach is to employ image enhancement or preprocessing before feature extraction. Super-resolution techniques, such as those used in \cite{park2023content,yue2023resshift}, and deblurring algorithms \cite{kong2023efficient,chen2024hierarchical}, can improve visual clarity and thereby aid recognition. However, these two-step pipelines introduce additional computational overhead and do not guarantee that the enhanced image features align optimally with those extracted from genuine high-quality images. Another approach is to retrain models using augmented datasets that incorporate degraded images, as explored in \cite{garg2020fast, ferrarini2022floppy, arcanjo2022droso}, but this often yields limited improvements due to the fundamental loss of visual information.

\subsection{Knowledge Distillation and Transfer Learning in VPR}
Likewise, research on effective methods to extract more representative global descriptors from low-quality images is lacking in the VPR community. Existing solutions have addressed the issue at the descriptor matching and image-retrieval level. The authors of \cite{tomita2023compress} propose a sequence-based image-retrieval strategy \cite{tomita2023sequence}, which queries the database images multiple times with lower per-query cost thanks to JPEG compression. Furthermore, predicting the retrieval success for each query image could favor points within the sequential queries that are more easily localized \cite{carson2023predict}. We instead wish to explore knowledge distillation as a means to directly improve descriptor extraction and thus increase the accuracy of database retrieval via individual queries.

After knowledge distillation was first proposed in \cite{hinton2015distilling}, it has been applied to various computer-vision tasks such as super-resolution \cite{hui2018fast}, image classification \cite{tung2019similarity,liu2021exploring}, and segmentation \cite{liu2021exploring}. More recently, knowledge distillation has also been used specifically in VPR to produce more discriminative global descriptors, such as learning from structural knowledge of an image-segmentation model \cite{shen2023structvpr}. We believe that knowledge distillation could similarly improve VPR accuracy for low-quality images.

%% file: Parts/3-Method.tex
\begin{figure}[t]
	\centering
	\includegraphics[width=1\columnwidth]{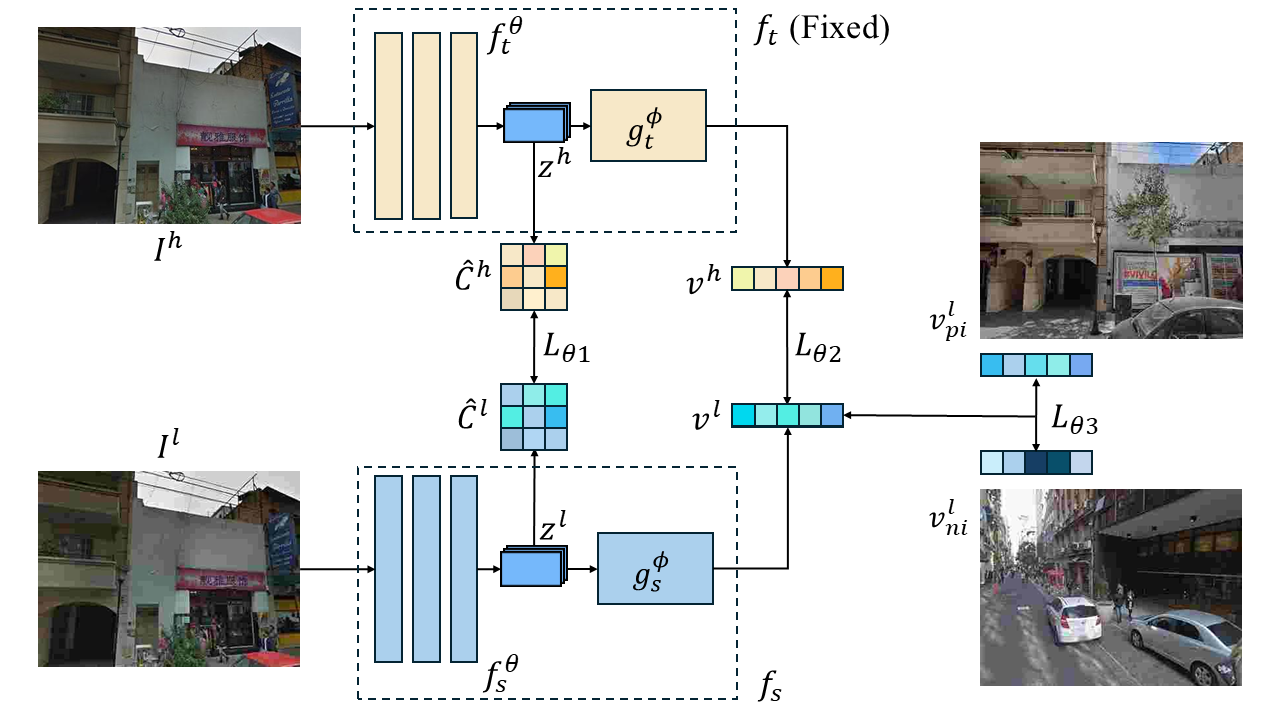}
    \caption{\textbf{Proposed Knowledge Distillation Methodology:} Through ICKD loss between latent codes \( z^h \) and \( z^l \), MSE loss between global descriptors \( v^h \) and \( v^l \), and triplet loss on \( v^l \), the student branch \( f_s \) learns the teacher branch \( f_t \)'s knowledge of high quality images \( I^h \) to extract more discriminative, representative \( v^l \) from low quality images \( I^l \).}
	\label{fig:distillation}
	\vspace{-15pt}
\end{figure}

\section{Distillation Methodology}

Our goal is to enhance existing VPR methods by enabling them to extract more representative global descriptors from low-quality images. To achieve this, we apply knowledge distillation techniques directly at the descriptor extraction level. Specifically, we enable a student descriptor extractor, which processes low-quality images, to approximate the outputs of a teacher network operating on high-quality images. This approach leverages the rich, discriminative information from high-quality images to compensate for the information loss in degraded inputs.

\subsection{Conceptual Model of Distillation for VPR}

Given some VPR method, we initialize two copies, designating one as the teacher branch \(f_t\) and the other as the student branch \(f_s\). The teacher processes a high-quality image \(I^h\) while the student processes its low-quality counterpart \(I^l\). As illustrated in Fig. \ref{fig:distillation}, the teacher model comprises a feature encoder \(f^{\theta}_t\) and a descriptor aggregator \(g^{\phi}_t\). Similarly, the student consists of a structurally identical encoder \(f^{\theta}_s\) and aggregator \(g^{\phi}_s\). Between the encoder and aggregator, the teacher and student produce intermediate 3D latent codes \(z^h\) and \(z^l\), respectively. These latent codes are subsequently aggregated into global descriptors \(v^h\) and \(v^l\). Our distillation strategy is built on three distinct loss functions designed to transfer the high-quality feature representation from the teacher to the student.

\subsection{ICKD Loss}
The Inter-Channel Correlation Knowledge Distillation (ICKD) loss \cite{liu2021exploring} is employed to ensure that the student network's latent code \(z^l\) closely mimics the teacher network's latent code \(z^h\) in a channel-wise manner. In Visual Place Recognition (VPR), each channel of the latent code can be viewed as a local feature descriptor, and it is critical that these features remain consistent regardless of image quality. By enforcing similarity in the inter-channel correlations between \(z^l\) and \(z^h\), we help preserve the discriminative structural information necessary for robust localization.

Owing to potential differences in spatial dimensions between \(I^l\) and \(I^h\), the latent codes \(z^l\) and \(z^h\) may also differ in size, making traditional distillation losses inapplicable. To overcome this, the ICKD loss computes an Inter-Channel Correlation (ICC) matrix from the latent codes, effectively reducing their spatial dimensions to a consistent \(c \times c\) form.

More specifically, let \(I^l\) have resolution \(W \times H\) and let the latent codes \(z^h\) and \(z^l\) have dimensions \(c \times W' \times H'\) and \(c \times w' \times h'\), respectively. By concatenating the spatial dimensions \(w'\) and \(h'\) for \(z^l\), we form a vector \(p^l\) of size \(c \times (w' \cdot h')\). This vector is normalized along its second dimension to yield \(\hat{p^l} = \frac{p^l}{\|p^l\|}\). The ICC matrix is then computed as \(C^l = \hat{p^l} \hat{p^l}^T\) (with dimensions \(c \times c\)) and normalized to obtain \(\hat{C^l} = \frac{C^l}{\|C^l\|}\). A similar process applied to \(z^h\) results in \(\hat{C^h}\). The ICKD loss is formulated as:
\begin{equation}
L_{\theta 1} = \|\hat{C^l} - \hat{C^h}\|_2.
\label{eqa: ICKD_loss}
\end{equation}
ensuring that the student encoder learns to mimic the channel correlations of the teacher, thereby preserving discriminative structural information despite spatial resolution differences.

\subsection{Mean Squared Error (MSE) Loss}
To directly align the global descriptors produced by the teacher and student branches, we employ a Mean Squared Error (MSE) loss. This loss minimizes the difference between \(v^l\) and \(v^h\), enforcing that the student produces descriptors similar to those of the teacher despite the quality degradation of its input:
\begin{equation}
L_{\theta 2} = \|v^l - v^h\|_2^2.
\label{eqa: MSE_loss}
\end{equation}

This term provides a straightforward supervision signal, encouraging the student to closely emulate the teacher's output.

\subsection{Weakly Supervised Triplet Ranking Loss}
The weakly supervised Triplet Ranking Loss is applied solely within the student branch \(f_s\) to further enhance the discriminative power of the learned descriptors. In our framework, following the same strategy as NetVLAD \cite{arandjelovic2016netvlad}, positive and negative samples are defined based on geographic proximity: images captured within a predefined distance (e.g., 25 meters) of the query are treated as positives, while those farther away are considered negatives. Inspired by this framework, our loss ensures that the descriptor \(v^l\) of a query image is closer (in Euclidean distance) to that of a geographically proximate (positive) image \(v^l_p\) than to that of a distant (negative) image \(v^l_n\). Formally, the loss enforces:
\[
d_\theta(v^l,v^l_p) < d_\theta(v^l,v^l_n),
\]
where \(d_\theta\) denotes the Euclidean distance.

The triplet loss is expressed as:
\begin{equation}
L_{\theta 3} = \sum_j l\left(\min_i d_\theta^2(v^l,{v^l_p}_i) - d_\theta^2(v^l,{v^l_n}_j) + m\right),
\label{eqa: Triplet_loss}
\end{equation}
with \(l(x) = \max(x, 0)\) being the hinge loss and \(m\) a margin parameter that ensures a sufficient gap between positive and negative pairs. Here, the term \(\min_i d_\theta^2(v^l,{v^l_p}_i)\) selects, among a set of candidate positives in the batch, the descriptor that is closest to the query. This selection is performed based on the current descriptor distances, which serves as a weak supervision signal because it relies on geographic proximity and descriptor similarity rather than explicit ground-truth labels. Although this strategy may occasionally select a false positive, the margin \(m\) and continuous network updates help mitigate this risk by ensuring that the model learns to focus on the most challenging and truly similar examples.

\subsection{Composite Loss}
Finally, the overall distillation objective is defined as a composite loss combining the three individual terms:
\begin{equation}
L = L_{\theta 1} + \alpha L_{\theta 2} + \beta L_{\theta 3},
\label{eqa:loss}
\end{equation}
where \(\alpha\) and \(\beta\) are weighting coefficients that balance the influence of the MSE and Triplet losses relative to the ICKD loss. In our experiments, we analyze both the individual and combined effects of these losses to validate the efficacy of our distillation approach.

% However, our experiments revealed that the triplet loss \(L_{\theta_3}\) actually hindered network performance, so our final loss becomes:
% \begin{equation}
% L = L_{\theta_1} + \alpha L_{\theta_2}.
% \label{eqa:loss_final}
% \end{equation}

% We now turn our attention to the examination of the proposed method.

%% file: Parts/4-Experiment.tex
\section{Experiments}
To evaluate the effectiveness of our distillation methodology, we analyze the impact of each loss component (\( L_{\theta1} \), \( L_{\theta2} \), \( L_{\theta3} \)) as well as their composite form \( L \) on VPR performance. We perform distillation on the student branch \( f_s \) using the GSV-Cities VPR dataset \cite{ali2022gsv} and subsequently evaluate retrieval accuracy on several benchmark datasets from the Deep Visual Geo-localization Benchmark \cite{berton2022bench}. For all experiments, low-quality images \( I^l \) are generated by applying 90\% JPEG compression (quality level = 10) to the original high-quality images \( I^h \). This choice allows us to simulate bandwidth-constrained scenarios while keeping the degradation consistent across training and testing.

Following the protocol in \cite{tomita2023sequence}, we evaluate \( f_s \) using a database and query set that both consist of \( I^l \) images. This controlled setting avoids inconsistencies due to variable compression levels and also illustrates the reduced cost of database construction when operating under strict quality constraints. Although our current experiments focus on JPEG compression, we further explore the generalizability of our method to low image resolution and high video quantization in Section \ref{extrares}.

\subsection{Datasets Configurations}

\subsubsection{Training Dataset} \label{trainset}
We use the GSV-Cities dataset, which contains approximately 530,000 street view images from 23 cities. To reduce training cost and assess our method under constrained data conditions, we use only the low-resolution subset (\(400\times300\)), reducing the training set to about 194,000 images.

\subsubsection{Testing Datasets} \label{testsets}
We evaluate \( f_s \) on three standard VPR datasets: Mapillary SLS\cite{warburg2020msls}, Nordland\cite{sunderhauf2013nordland}, and Tokyo 24/7\cite{torii2018tokyo247}

\setlength{\tabcolsep}{5.5pt}
\input{Table_Figure_Wrappers/table2}

\subsection{VPR Methods}
To validate the broad applicability of our knowledge distillation framework, we experiment with five VPR methods representing different architectures:
\begin{itemize}
    \item NetVLAD \cite{arandjelovic2016netvlad} and MixVPR \cite{ali2023mixvpr} (CNN-based),
    \item AnyLoc \cite{keetha2023anyloc}, DINOv2SALAD \cite{izquierdo2024salad}, and CricaVPR \cite{lu2024cricavpr} (DINOv2-based \cite{oquab2024dinov2}).
\end{itemize}
For each method, we initialize both the teacher \( f_t \) and student \( f_s \) using pretrained weights released by each method's authors for faster convergence. To investigate the effects of individual loss components, we train \( f_s \) under all seven possible combinations of \( L_{\theta1} \), \( L_{\theta2} \), and \( L_{\theta3} \) (with at least one term used from Eq. (\ref{eqa:loss})). 

As a baseline, we also fine-tune each method (except AnyLoc, which uses a frozen DINOv2 backbone) on the same \( I^l \) images using pretrained weights and the authors' provided training procedure instead of our distillation losses. This baseline comparison highlights the improvement attributable solely to our distillation framework.

\subsection{Experimental Setup}

\subsubsection{Training Configuration} \label{trainconf}
Our loss weighting coefficients in Eq. (\ref{eqa:loss}) are set to \(\alpha=10^5\) and \(\beta=10^4\). These values were chosen based on preliminary ablation studies aimed at balancing the scale of the MSE and triplet losses relative to the ICKD loss. The learning rate is initialized at \( 10^{-5} \) and adjusted using a decay factor of \( 2\times10^{-11} \) along with an exponential decay rate of \( 0.99999 \). Each training experiment (one of the seven loss combinations applied to one of the five VPR methods) comprises a single epoch over the reduced GSV-Cities dataset (Section \ref{trainset}). For the triplet loss (Eq. (\ref{eqa: Triplet_loss})), we sample 5 negative images per training sample to form \( v^l_n \).

\subsubsection{VPR Performance Evaluation} \label{recalldef}
We assess the VPR performance of the distilled student network \( f_s \) using the Recall at N (R@N) metric, following prior work \cite{arandjelovic2016netvlad,ali2023mixvpr,hausler2021patch,berton2022rethinking,izquierdo2024salad,keetha2023anyloc,lu2024cricavpr}. R@N is defined as the percentage of queries for which at least one of the top \( N \) retrieved database images is within a specified distance threshold \( d \) of the query's true location. Following the VPR methods we experimented with, we set \( d \) to $25$ meters, and we evaluate R@N for \( N \) values of 1, 2, 5, and 10.

%% file: Table_Figure_Wrappers/table2.tex
% Table generated by Excel2LaTeX from sheet 'table II'
\begin{table*}[t]
  \centering
  \caption{\textbf{Best Performance of VPR Methods after Distillation:} For low quality images, the loss combination producing the highest recall rate is compared against the baseline of fine-tuning. For both distilled and fine-tuned weights, the change in VPR recall is represented as a delta relative to the performance of each method using pretrained weights. Within each dataset and method, \textcolor[rgb]{ .329,  .51,  .208}{greene} text indicates the greatest improvement for every R@N, whereas any decrease relative to pretrained performance is marked as \textcolor[rgb]{ .612,  0,  .024}{red}. The recall rates using pretrained weights on unmodified $I^h$ images are provided as reference.}
    \begin{tabular}{cr|rrrr|rrrr|rrrr}
    \toprule
    \multirow{2}{*}{VPR Methods} & \multicolumn{1}{c|}{\multirow{2}{*}{Configuration}} & \multicolumn{4}{c|}{Mapillary SLS} & \multicolumn{4}{c|}{Nordland} & \multicolumn{4}{c}{Tokyo 24/7} \\
          &       & \multicolumn{1}{l}{R@1} & \multicolumn{1}{l}{R@2} & \multicolumn{1}{l}{R@5} & \multicolumn{1}{l|}{R@10} & \multicolumn{1}{l}{R@1} & \multicolumn{1}{l}{R@2} & \multicolumn{1}{l}{R@5} & \multicolumn{1}{l|}{R@10} & \multicolumn{1}{l}{R@1} & \multicolumn{1}{l}{R@2} & \multicolumn{1}{l}{R@5} & \multicolumn{1}{l}{R@10} \\
    \midrule
    \multirow{4}{*}{MixVPR} & \textcolor[rgb]{ .651,  .651,  .651}{pretrained ($I^h$)} & \textcolor[rgb]{ .651,  .651,  .651}{82.73} & \textcolor[rgb]{ .651,  .651,  .651}{86.67} & \textcolor[rgb]{ .651,  .651,  .651}{89.73} & \textcolor[rgb]{ .651,  .651,  .651}{91.65} & \textcolor[rgb]{ .651,  .651,  .651}{57.79} & \textcolor[rgb]{ .651,  .651,  .651}{64.13} & \textcolor[rgb]{ .651,  .651,  .651}{71.49} & \textcolor[rgb]{ .651,  .651,  .651}{76.41} & \textcolor[rgb]{ .651,  .651,  .651}{87.30} & \textcolor[rgb]{ .651,  .651,  .651}{89.52} & \textcolor[rgb]{ .651,  .651,  .651}{92.06} & \textcolor[rgb]{ .651,  .651,  .651}{93.65} \\
          & pretrained ($I^l$)& 71.87 & 76.61 & 81.22 & 84.19 & 31.05 & 36.12 & 44.13 & 51.23 & 66.03 & 73.33 & 78.73 & 82.22 \\
          & finetuned & \textcolor[rgb]{ .329,  .51,  .208}{+4.43} & \textcolor[rgb]{ .329,  .51,  .208}{+3.86} & \textcolor[rgb]{ .329,  .51,  .208}{+3.60} & \textcolor[rgb]{ .329,  .51,  .208}{+2.99} & +13.37 & +14.28 & +14.64 & +14.46 & \textcolor[rgb]{ .329,  .51,  .208}{+9.52} & \textcolor[rgb]{ .329,  .51,  .208}{+7.62} & +6.03 & +5.40 \\
          & \textbf{ICKD} & \textbf{+4.33} & \textbf{+3.69} & \textbf{+3.20} & \textbf{+2.93} & \textcolor[rgb]{ .329,  .51,  .208}{\textbf{+15.00}} & \textcolor[rgb]{ .329,  .51,  .208}{\textbf{+16.59}} & \textcolor[rgb]{ .329,  .51,  .208}{\textbf{+16.78}} & \textcolor[rgb]{ .329,  .51,  .208}{\textbf{+15.94}} & \textbf{+8.25} & \textbf{+6.03} & \textcolor[rgb]{ .329,  .51,  .208}{\textbf{+6.35}} & \textcolor[rgb]{ .329,  .51,  .208}{\textbf{+6.67}} \\
    \midrule
    \multirow{4}{*}{CricaVPR} & \textcolor[rgb]{ .651,  .651,  .651}{pretrained ($I^h$)} & \textcolor[rgb]{ .651,  .651,  .651}{74.74} & \textcolor[rgb]{ .651,  .651,  .651}{80.92} & \textcolor[rgb]{ .651,  .651,  .651}{86.09} & \textcolor[rgb]{ .651,  .651,  .651}{88.75} & \textcolor[rgb]{ .651,  .651,  .651}{87.64} & \textcolor[rgb]{ .651,  .651,  .651}{90.65} & \textcolor[rgb]{ .651,  .651,  .651}{94.24} & \textcolor[rgb]{ .651,  .651,  .651}{95.80} & \textcolor[rgb]{ .651,  .651,  .651}{90.16} & \textcolor[rgb]{ .651,  .651,  .651}{92.38} & \textcolor[rgb]{ .651,  .651,  .651}{95.56} & \textcolor[rgb]{ .651,  .651,  .651}{96.19} \\
          & pretrained ($I^l$)& 68.14 & 74.41 & 80.24 & 83.19 & 63.51 & 69.71 & 77.46 & 82.64 & 74.29 & 79.68 & 84.76 & 87.94 \\
          & finetuned & \textcolor[rgb]{ .612,  0,  .024}{-0.12} & \textcolor[rgb]{ .612,  0,  .024}{-0.64} & \textcolor[rgb]{ .612,  0,  .024}{-0.77} & \textcolor[rgb]{ .612,  0,  .024}{-0.23} & \textcolor[rgb]{ .612,  0,  .024}{-2.97} & \textcolor[rgb]{ .612,  0,  .024}{-2.93} & \textcolor[rgb]{ .612,  0,  .024}{-2.97} & \textcolor[rgb]{ .612,  0,  .024}{-2.79} & +5.08 & +4.44 & +2.54 & +1.90 \\
          & \textbf{ICKD} & \textcolor[rgb]{ .329,  .51,  .208}{\textbf{+0.89}} & \textcolor[rgb]{ .329,  .51,  .208}{\textbf{+1.39}} & \textcolor[rgb]{ .329,  .51,  .208}{\textbf{+1.11}} & \textcolor[rgb]{ .329,  .51,  .208}{\textbf{+1.14}} & \textcolor[rgb]{ .329,  .51,  .208}{\textbf{+11.63}} & \textcolor[rgb]{ .329,  .51,  .208}{\textbf{+10.80}} & \textcolor[rgb]{ .329,  .51,  .208}{\textbf{+8.95}} & \textcolor[rgb]{ .329,  .51,  .208}{\textbf{+7.03}} & \textcolor[rgb]{ .329,  .51,  .208}{\textbf{+6.98}} & \textcolor[rgb]{ .329,  .51,  .208}{\textbf{+5.08}} & \textcolor[rgb]{ .329,  .51,  .208}{\textbf{+6.03}} & \textcolor[rgb]{ .329,  .51,  .208}{\textbf{+4.76}} \\
    \midrule
    \multirow{4}{*}{\makecell{DINOv2\\SALAD}} & \textcolor[rgb]{ .651,  .651,  .651}{pretrained ($I^h$)} & \textcolor[rgb]{ .651,  .651,  .651}{89.20} & \textcolor[rgb]{ .651,  .651,  .651}{92.40} & \textcolor[rgb]{ .651,  .651,  .651}{94.70} & \textcolor[rgb]{ .651,  .651,  .651}{95.84} & \textcolor[rgb]{ .651,  .651,  .651}{88.08} & \textcolor[rgb]{ .651,  .651,  .651}{90.94} & \textcolor[rgb]{ .651,  .651,  .651}{94.13} & \textcolor[rgb]{ .651,  .651,  .651}{95.98} & \textcolor[rgb]{ .651,  .651,  .651}{97.14} & \textcolor[rgb]{ .651,  .651,  .651}{97.46} & \textcolor[rgb]{ .651,  .651,  .651}{98.73} & \textcolor[rgb]{ .651,  .651,  .651}{99.05} \\
          & pretrained ($I^l$)& 84.60 & 88.85 & 91.89 & 93.68 & 67.90 & 73.88 & 80.40 & 84.24 & 89.21 & 92.06 & 95.87 & 96.51 \\
          & finetuned & \textcolor[rgb]{ .612,  0,  .024}{-0.22} & \textcolor[rgb]{ .612,  0,  .024}{-0.57} & \textcolor[rgb]{ .612,  0,  .024}{-0.60} & \textcolor[rgb]{ .612,  0,  .024}{-0.76} & +0.04 & \textcolor[rgb]{ .612,  0,  .024}{-0.40} & \textcolor[rgb]{ .612,  0,  .024}{-0.76} & \textcolor[rgb]{ .612,  0,  .024}{-0.04} & \textcolor[rgb]{ .612,  0,  .024}{-0.63} & 0.00  & \textcolor[rgb]{ .612,  0,  .024}{-0.95} & 0.00 \\
          & \textbf{ICKD} & \textcolor[rgb]{ .329,  .51,  .208}{\textbf{+0.52}} & \textcolor[rgb]{ .329,  .51,  .208}{\textbf{+0.45}} & \textcolor[rgb]{ .329,  .51,  .208}{\textbf{+0.20}} & \textcolor[rgb]{ .329,  .51,  .208}{\textbf{+0.30}} & \textcolor[rgb]{ .329,  .51,  .208}{\textbf{+1.56}} & \textcolor[rgb]{ .329,  .51,  .208}{\textbf{+1.81}} & \textcolor[rgb]{ .329,  .51,  .208}{\textbf{+1.09}} & \textcolor[rgb]{ .329,  .51,  .208}{\textbf{+1.20}} & \textcolor[rgb]{ .329,  .51,  .208}{\textbf{+1.59}} & \textcolor[rgb]{ .329,  .51,  .208}{\textbf{+1.27}} & \textcolor[rgb]{ .329,  .51,  .208}{\textbf{+0.32}} & 0.00 \\
    \midrule
    \multirow{4}{*}{NetVLAD} & \textcolor[rgb]{ .651,  .651,  .651}{pretrained ($I^h$)} & \textcolor[rgb]{ .651,  .651,  .651}{49.29} & \textcolor[rgb]{ .651,  .651,  .651}{55.24} & \textcolor[rgb]{ .651,  .651,  .651}{62.07} & \textcolor[rgb]{ .651,  .651,  .651}{67.29} & \textcolor[rgb]{ .651,  .651,  .651}{5.51} & \textcolor[rgb]{ .651,  .651,  .651}{6.67} & \textcolor[rgb]{ .651,  .651,  .651}{8.62} & \textcolor[rgb]{ .651,  .651,  .651}{11.38} & \textcolor[rgb]{ .651,  .651,  .651}{60.63} & \textcolor[rgb]{ .651,  .651,  .651}{63.81} & \textcolor[rgb]{ .651,  .651,  .651}{69.21} & \textcolor[rgb]{ .651,  .651,  .651}{74.29} \\
          & pretrained ($I^l$)& 32.60 & 37.67 & 44.92 & 50.13 & 1.99  & 3.01  & 4.93  & 7.07  & 27.94 & 33.02 & 41.90 & 47.94 \\
          & finetuned & +0.03 & +0.01 & \textcolor[rgb]{ .612,  0,  .024}{-0.02} & \textcolor[rgb]{ .612,  0,  .024}{-0.01} & 0.00  & 0.00  & 0.00  & 0.00  & 0.00  & 0.00  & 0.00  & 0.00 \\
          & \textbf{MSE} & \textcolor[rgb]{ .329,  .51,  .208}{\textbf{+4.46}} & \textcolor[rgb]{ .329,  .51,  .208}{\textbf{+5.32}} & \textcolor[rgb]{ .329,  .51,  .208}{\textbf{+5.09}} & \textcolor[rgb]{ .329,  .51,  .208}{\textbf{+5.01}} & \textcolor[rgb]{ .329,  .51,  .208}{\textbf{+0.54}} & \textcolor[rgb]{ .329,  .51,  .208}{\textbf{+0.43}} & \textcolor[rgb]{ .612,  0,  .024}{-0.18} & \textcolor[rgb]{ .612,  0,  .024}{-0.72} & \textcolor[rgb]{ .329,  .51,  .208}{\textbf{+8.25}} & \textcolor[rgb]{ .329,  .51,  .208}{\textbf{+6.35}} & \textcolor[rgb]{ .329,  .51,  .208}{\textbf{+6.03}} & \textcolor[rgb]{ .329,  .51,  .208}{\textbf{+5.08}} \\
    \midrule
    \multirow{3}{*}{AnyLoc} & \textcolor[rgb]{ .651,  .651,  .651}{pretrained ($I^h$)} & \textcolor[rgb]{ .651,  .651,  .651}{56.51} & \textcolor[rgb]{ .651,  .651,  .651}{62.31} & \textcolor[rgb]{ .651,  .651,  .651}{68.28} & \textcolor[rgb]{ .651,  .651,  .651}{72.89} & \textcolor[rgb]{ .651,  .651,  .651}{12.90} & \textcolor[rgb]{ .651,  .651,  .651}{16.23} & \textcolor[rgb]{ .651,  .651,  .651}{20.18} & \textcolor[rgb]{ .651,  .651,  .651}{24.28} & \textcolor[rgb]{ .651,  .651,  .651}{88.25} & \textcolor[rgb]{ .651,  .651,  .651}{91.11} & \textcolor[rgb]{ .651,  .651,  .651}{94.92} & \textcolor[rgb]{ .651,  .651,  .651}{96.83} \\
          & pretrained ($I^l$)& 48.04 & 56.45 & 63.84 & 69.17 & 10.11 & 12.97 & 17.86 & 21.92 & 83.49 & 87.30 & 91.75 & 95.87 \\
          & \textbf{ICKD + Triplet} & \textcolor[rgb]{ .329,  .51,  .208}{\textbf{+0.99}} & \textcolor[rgb]{ .329,  .51,  .208}{\textbf{+1.54}} & \textcolor[rgb]{ .329,  .51,  .208}{\textbf{+2.68}} & \textcolor[rgb]{ .329,  .51,  .208}{\textbf{+2.44}} & \textcolor[rgb]{ .329,  .51,  .208}{\textbf{+1.74}} & \textcolor[rgb]{ .329,  .51,  .208}{\textbf{+1.88}} & \textcolor[rgb]{ .329,  .51,  .208}{\textbf{+1.52}} & \textcolor[rgb]{ .329,  .51,  .208}{\textbf{+2.54}} & \textcolor[rgb]{ .612,  0,  .024}{-5.71} & \textcolor[rgb]{ .612,  0,  .024}{-1.59} & \textcolor[rgb]{ .329,  .51,  .208}{\textbf{+1.59}} & \textcolor[rgb]{ .612,  0,  .024}{-0.95} \\
    \bottomrule
    \end{tabular}%
  \label{table:mainres_test}%
\vspace{-10pt}
\end{table*}%

%% file: Parts/5-Results.tex
\begin{figure*}[t]
        \centering
	\includegraphics[width=1\textwidth]{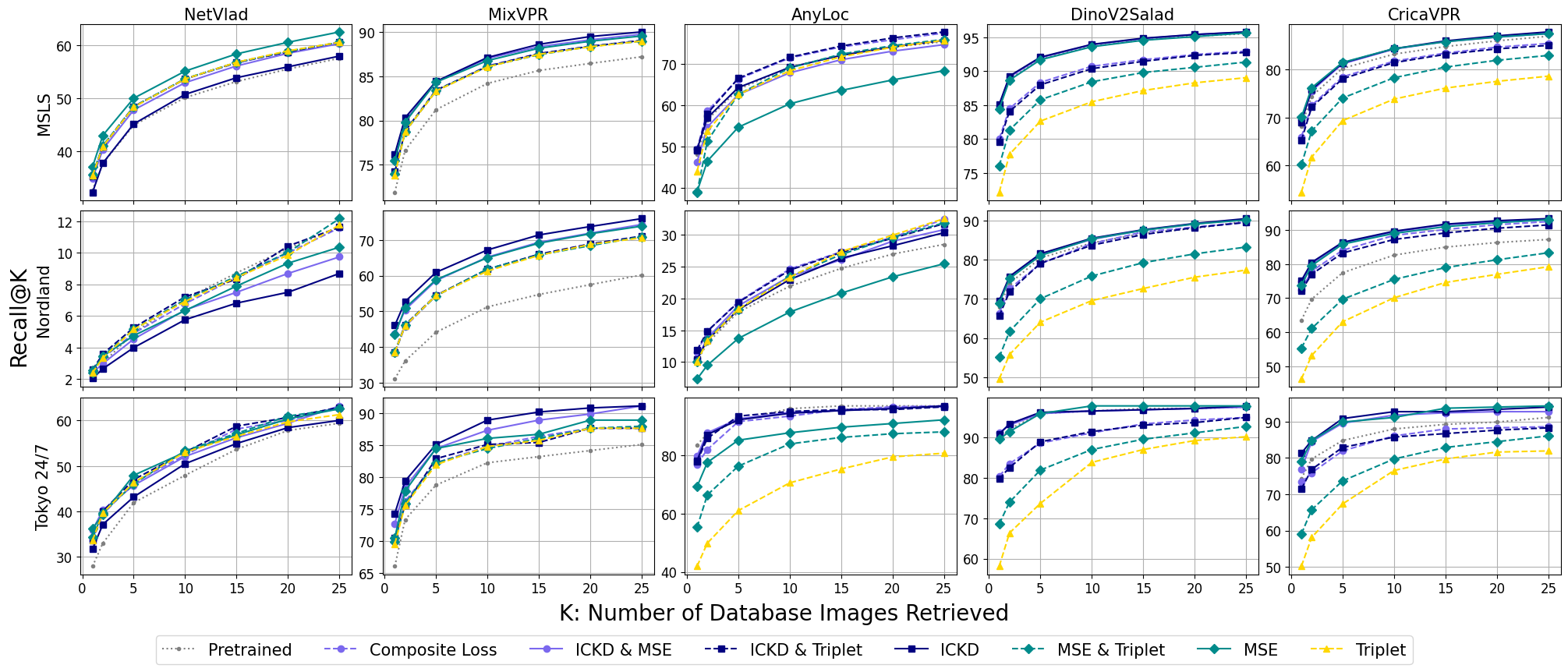}
    \captionof{figure}{\textbf{Recall Rates for All Losses:} For each method, we compare \( f_s \) after distillation with loss combinations as well as pretrained weights. Different marker shapes indicate different combinations of ICKD and MSE losses, whereas including the triplet loss is shown with a dashed line. For readability, the y-axis of each subplot is independently scaled.}
	\label{fig:loss_combo}
    \vspace{-10pt}
\end{figure*}

\section{Results and Discussions}
We analyze the performance of the student network \(f_s\) after distillation using each loss combination. We report quantitative improvements in VPR recall rates, discuss observed trends across different methods and datasets, and provide qualitative visualizations to interpret the behavior of the network after distillation.

%the current subsubsections I have put makes this subsection more like ablation study, but the distillation framework is strong when isolated
\subsection{Quantitative Results}
Table \ref{table:mainres_test} demonstrates the applicability of (\ref{eqa:loss}) and its components to a representative selection of VPR methods and datasets. Albeit with varying magnitudes, the deltas of recall rates under \( v^l \) after distillation and those under \( v^l \) from pretrained weights are generally positive. However, fine-tuning is inconsistent and mostly less effective than distillation. The Nordland recall rates for MixVPR and CricaVPR, two architecturally distinct methods, exemplifies distillation's superior efficacy and applicability. There are only two instances where distillation partially failed to increase VPR recall rates under \( v^l \) for some method and dataset, namely NetVLAD on Nordland and AnyLoc on Tokyo 24/7. As AnyLoc's recall rates fluctuate greatly on Tokyo 24/7, we postulate the combination of the natures of this method and dataset to be an edge case. Since NetVLAD's improvement on Nordland for all R@N is less than that on other datasets, we reserve this phenomenon for further analysis in \ref{qualitative}.

Besides numerically demonstrating the most effective distillation losses, we graphically compare each loss combination in fig. \ref{fig:loss_combo}, where we analyze full recall rate data for all variants of \( f_s \). Previously, we have observed that ICKD (\ref{eqa: ICKD_loss}) produces best VPR performances for MixVPR and CricaVPR and that MSE (\ref{eqa: MSE_loss}) is best for NetVLAD, and these three methods benefit the most from distillation. Now, the relative superiority of the two aforementioned losses could be seen on all methods, where the recall rates for \( f_s \) trained under either or both of them generally trend towards the top of all loss combinations. Here, AnyLoc and NetVLAD on Nordland are exceptions, where augmenting (\ref{eqa: ICKD_loss}) or (\ref{eqa: MSE_loss}) with triplet loss (\ref{eqa: Triplet_loss}) produces a better-performing \( f_s \). However, besides NetVLAD, whose original training scheme uses the triplet loss, (\ref{eqa: Triplet_loss}) is generally the least effective when used alone.

% Table generated by Excel2LaTeX from sheet 'table III'
\setlength{\tabcolsep}{5pt}
\begin{table}[t]
  \centering
  \caption{Indoor Evaluation Using Pretrained Weights}
    \begin{tabular}{rr|rr|rr|rr}
    \toprule
    \multicolumn{1}{c}{\multirow{2}{*}{Methods}} & \multicolumn{1}{c|}{\multirow{2}{*}{Quality}} & \multicolumn{2}{c|}{TUM LSI} & \multicolumn{2}{c|}{\makecell{Gangnam\\Station}} & \multicolumn{2}{c}{\makecell{NYC-Indoor-\\VPR}} \\
          &       & R@1   & R@5   & R@1   & R@5   & R@1   & R@5 \\
    \midrule
    \multirow{2}{*}{MixVPR} & $I^h$   & 94.09 & 99.55 & 4.39  & 13.63 & 41.23 & 83.76 \\
          & 90\%  & 91.36 & 99.09 & 3.93  & 11.76 & 40.04 & 80.76 \\
    \midrule
    \multirow{2}{*}{CricaVPR} & $I^h$   & 92.27 & 99.09 & 7.48  & 21.64 & 37.92 & 82.36 \\
          & 90\%  & 92.27 & 97.27 & 7.67  & 21.37 & 37.30 & 79.26 \\
    \midrule
    \multirow{2}{*}{\makecell{DINOv2\\SALAD}} & $I^h$   & 94.09 & 99.09 & 8.93  & 22.67 & 39.89 & 82.98 \\
          & 90\%  & 91.82 & 98.64 & 9.43  & 21.37 & 39.58 & 81.69 \\
    \midrule
    \multirow{2}{*}{NetVLAD} & $I^h$   & 95.00 & 99.09 & 3.21  & 8.36  & 39.42 & 81.27 \\
          & 90\%  & 91.82 & 98.18 & 2.94  & 9.81  & 38.23 & 79.10 \\
    \midrule
    \multirow{2}{*}{AnyLoc} & $I^h$   & 97.73 & 99.55 & 4.47  & 12.79 & 37.92 & 82.77 \\
          & 90\%  & 93.64 & 98.18 & 3.63  & 11.11 & 36.37 & 81.07 \\
    \bottomrule
    \end{tabular}%
  \label{table:indoorpre}%
  \vspace{-12pt}
\end{table}%

\subsection{Qualitative results} \label{qualitative}
\begin{figure*}[t]
    \centering
    \subfloat{\includegraphics[width=0.32\textwidth]{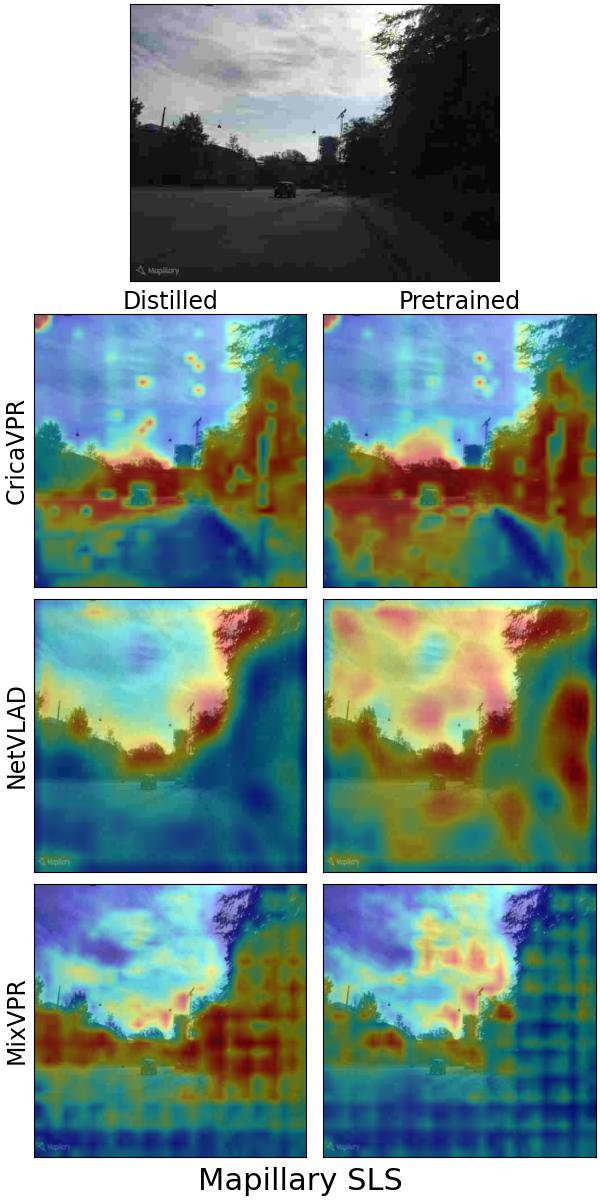}}
    \subfloat{\includegraphics[width=0.32\textwidth]{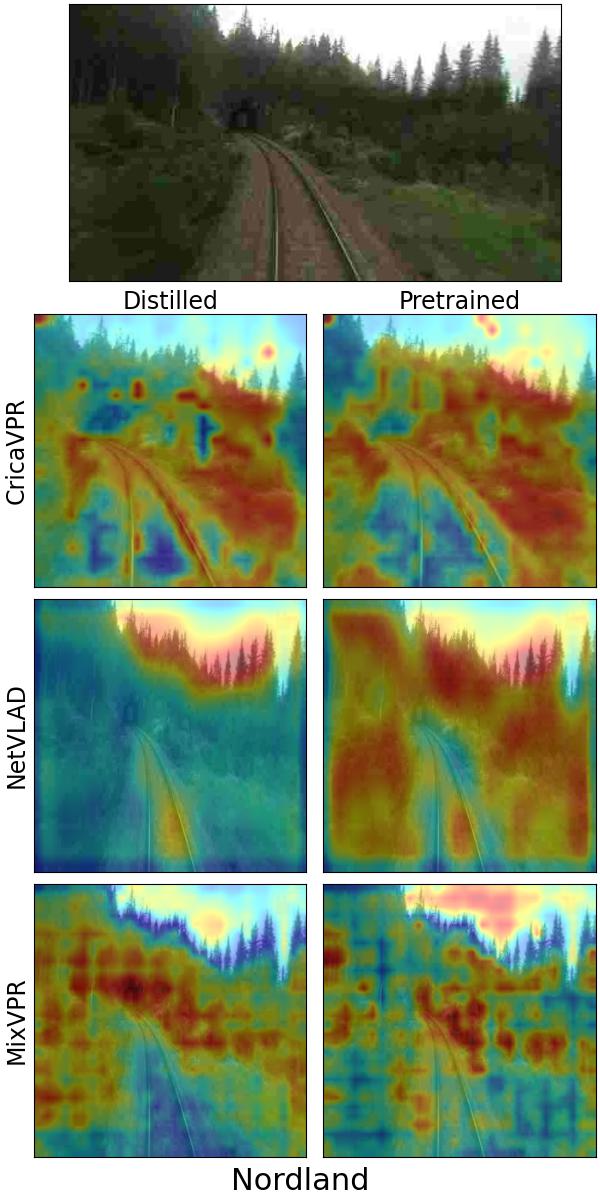}}
    \subfloat{\includegraphics[width=0.32\textwidth]{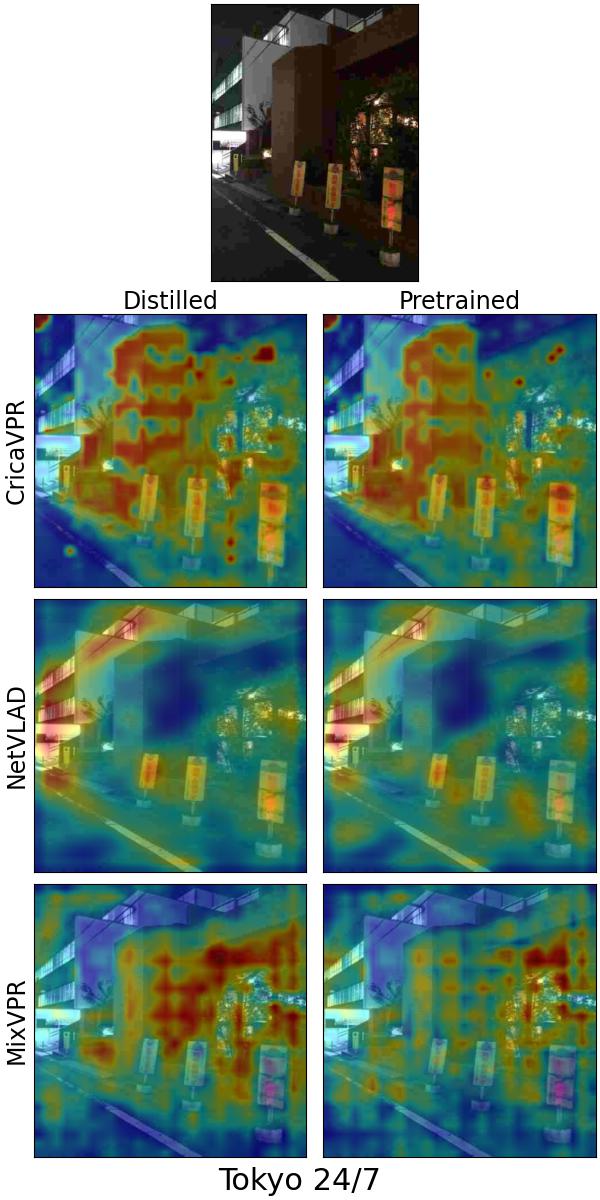}}
    \caption{\textbf{Activation Maps:} The selected \( I^l \) queries (top of figure) are successfully recalled by \( f_s \), while \( f_t \) fails. For CricaVPR, the focus of its ICKD-trained \( f_s \) is visualized by averaging its \( z^l \) across the channel dimension in accordance with its authors. For NetVLAD, we averaged a weighted sum of its MSE-trained \( z^l \) via tentative cluster assignments from \( g^{\phi}_s \). For MixVPR, since its authors did not visualize activations, we masked small regions of \( I^l \) and considered the resulting change in its ICKD-trained \( v_l \) as activations, creating block-like patches on its heatmap.}
	\label{fig:temp_heatmap}
    \vspace{-10pt}
\end{figure*}

We further investigate distillation's effects on MixVPR, CricaVPR, and NetVLAD, the most improved methods, and potentially account for the latter's more limited improvement on Nordland. In fig. \ref{fig:temp_heatmap}, we plot activation maps for each method's encoder \( f^{\theta}_s \) after distillation with their respective best-performing loss. From each of the three VPR testing datasets \ref{testsets}, we visualize an \( I^l \) query where \( f_s \) improves upon \( f_t \). On Mapillary SLS and Nordland, distillation mitigates the distraction of uninformative sky features especially for NetVLAD and MixVPR. More generally, distillation shifts focus from repetitive foreground features, such as road surfaces and ground-level vegetation, to more structurally distinct whole-scene features, such as the contours of buildings and farther tree lines. In the case of NetVLAD, the highly selective focus of \( f^{\theta}_s \) could account for its lesser recall rate improvement on Nordland. For the other two methods, while MixVPR's improvement could be more intuitively attributed to increased overall focus on informative image regions, CricaVPR's comparatively smaller change in focus suggests that its improvements originates from other factors, possibly concerning its aggregator \( g^{\phi}_s \).

% \begin{figure*}[t]
% 	\centering
% 	\includegraphics[width=1\textwidth]{Figures/figure3_loss_combos.png}
%     \caption{R@1 for all loss combinations. \textbf{I do not intend to simultaneously keep this figure and \ref{fig:loss_combo}, just illustrating the info density point from earlier} . In some ways the bar chart is more readable, but it's also less space-efficient, and the reader can't identify each loss combination as well (of course I can also make a legend for a bar chart and increase the font size, but I'm not sure if the problem goes away). Here blue means improvement, and red means deterioration.}
% 	\label{fig:loss_bar_top1}
% \end{figure*}

%% file: Parts/6-AdditionalExperiments.tex
\section{Additional Experiments} \label{extrares}
While our knowledge distillation approach already demonstrates notable efficacy for the majority of methods and datasets tested, we extend distillation to more datasets and other modes of image quality reduction discussed in \ref{low_quality_ways}.

\subsection{JPEG Compression's Impacts on Indoor Datasets}
While our previous VPR testing datasets cover both rural and urban environments and contain temporal changes and perceptual aliasing, they all capture outdoor environments. In table \ref{table:indoorpre}, We further examine the impacts of JPEG compression on three indoor datasets of different difficulty levels, namely TUM LSI \cite{walch2017tumlsi}, Gangnam Station (one of NAVER LABS' large-scale localization datasets \cite{lee2021large}), and NYC-Indoor-VPR \cite{sheng2024nyc}. While the performance of all methods with pretrained weights fluctuate greatly across datasets, JPEG compression's impacts on recall rates are significantly smaller than previous results with outdoor datasets. We postulate that due to indoor environments generally lack large contiguous regions of uninformative features such as sky, the proportion of distinctive features degraded by JPEG compression is smaller within indoor images. Therefore, instead of repeating our previous experiments, we extend our distillation methodology to other possible forms of \( I^l \).

\subsection{Other Modalities of Image Quality Reduction}
As JPEG compression had greater impacts on outdoor scenes, we use outdoor datasets from \ref{testsets} to yield \( I^l \) with reduced resolution instead of JPEG compression. Secondly, we explore the influence of video quantization on VPR recall, mirroring realistic practice of streaming image data through video. However, the image-based nature of existing VPR datasets prompts us to curate a custom video-based VPR dataset to produce \( I^l \) with increased video quantization. 

Among the five methods in table \ref{table:mainres_test}, NetVLAD suffers the most from lowered image quality in general. Thus, we select NetVLAD as the candidate of verifying knowledge distillation's benefits for VPR performance under resolution and quantization-based \( I^l \).

\subsection{Custom Video-Based VPR Dataset} 
Our custom VPR dataset is sourced from indoor spaces of the 6th floor of the Lighthouse Guild, an eye care facility in New York City. Instead of following existing indoor VPR datasets in capturing images \cite{walch2017tumlsi,lee2021large,sheng2021nyu,sheng2024nyc}, we record panoramic videos using the Insta360 camera, providing a $360$\textdegree\hspace{1pt} view. Each frame, with dimensions \(5760 \times 2880\), was horizontally segmented into \(18\) perspective images with $90$\textdegree\hspace{1pt} field of view and dimensions \(1440 \times 810\). These images were assembled into perspective videos for both resolution reduction and video quantization according to the following:
\begin{itemize}
    \item Resolutions: $405$p (\(720 \times 405\)), $203$p (\(360 \times 203\))
    \item Quantization Parameters: \(30\), \(33\), \(36\), \(39\), \(42\), \(45\), \(48\)
\end{itemize}

\begin{figure*}[t]
	\centering
	\includegraphics[width=1\textwidth]{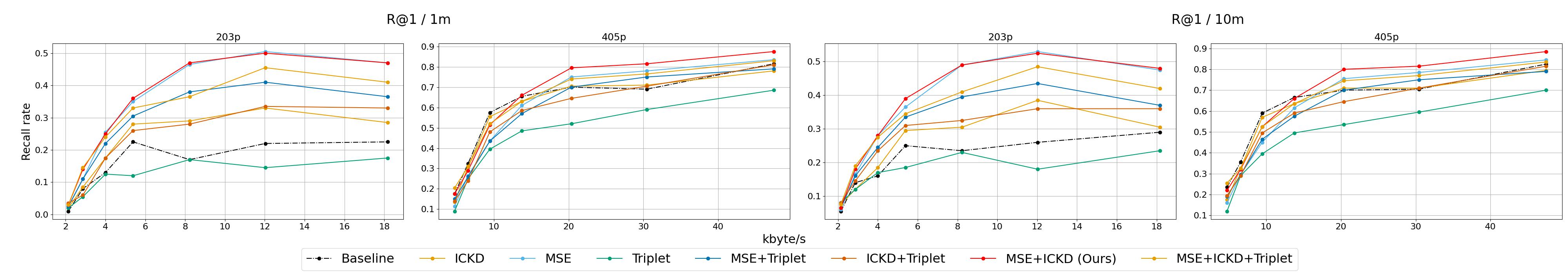}
    \caption{\textbf{Results on Our Dataset:} The recall rates of Pitts250k-trained \( f_s \) are obtained for each quantization parameter. For ease of visualization, the latter is converted into the video bitrate metric with lower bitrate indicating higher quantization. The left two plots were calculated with a VPR threshold of $1$m, and the right two have a $10$m threshold.}
	\label{fig:validation}
	\vspace{-12pt}
\end{figure*}

\subsection{Resolution and Quantization Experimental Results}
We not only validate distillation's efficacy for different forms of \( I^l \) but also its independence to datasets and \( I^l \)'s modality during training. While following \ref{trainconf}'s steps, we perform distillation on NetVLAD using the Pitts250k dataset \cite{torii2013visual}, which contains around $250,000$ outdoor $640\times480$ images of Pittsburgh's urban environments. All images are reduced to $240\times180$ as \( I^l \) for \( f_s \) to learn on.
To analyze distillation's robustness to changing \( I^l \) modalities, we test NetVLAD's recall on our custom dataset with \( v^l \) extracted from resolutions of $405$p and $203$p and quantization parameters spanning from $30$ to $48$ in intervals of $3$. For previous outdoor datasets, we extract \( v^l \) from $320\times240$ ($240$p). The influence of our dataset's various resolutions and quantization levels on \( f_s \)'s recall rate after distillation is compared against pretrained weights as a baseline in Fig. \ref{fig:validation}. The $203$p results restate MSE (\ref{eqa: MSE_loss})'s efficacy on NetVLAD, but despite ICKD (\ref{eqa: ICKD_loss})'s comparatively inferior performance on constant-modality \( I^l \) before, combining the two losses further improves VPR performance on $405$p, suggesting ICKD's greater adaptability under video quantization. Notably, changing recall distance thresholds as defined in \ref{recalldef} has little effect on VPR performance.

For outdoor datasets, we compare pretrained NetVLAD against trained \( f_s \) with the highest recall rate in table \ref{table:netvlad_lowres}. NetVLAD's best-performing distillation loss remains MSE, but adding ICKD yields stronger performance on Tokyo 24/7. This corroborates the previous observation of ICKD's adaptability, which may allow \( f_s \) to better withstand changing \( I^l \) modalities from $180$p during distillation to $240$p in testing.

% Table generated by Excel2LaTeX from sheet 'table IV'
\setlength{\tabcolsep}{5.5pt}
\begin{table}[t]
  \caption{NetVLAD's Performance on $240$p Images}
  \centering
    \begin{tabular}{cr|rrrr}
    \toprule
    \multicolumn{1}{l}{Dataset} & \multicolumn{1}{l|}{Configuration} & \multicolumn{1}{l}{R@1} & \multicolumn{1}{l}{R@2} & \multicolumn{1}{l}{R@5} & \multicolumn{1}{l}{R@10} \\
    \midrule
    \multirow{2}[2]{*}{Mapillary SLS} & pretrained & 31.16 & 37.89 & 46.36 & 52.54 \\
          & \textbf{MSE} & \textcolor[rgb]{ .329,  .51,  .208}{\textbf{+7.57}} & \textcolor[rgb]{ .329,  .51,  .208}{\textbf{+7.88}} & \textcolor[rgb]{ .329,  .51,  .208}{\textbf{+7.64}} & \textcolor[rgb]{ .329,  .51,  .208}{\textbf{+7.54}} \\
    \midrule
    \multirow{2}[2]{*}{Nordland} & pretrained & 3.99  & 5.40  & 8.34  & 10.40 \\
          & \textbf{MSE} & \textcolor[rgb]{ .329,  .51,  .208}{\textbf{+1.12}} & \textcolor[rgb]{ .329,  .51,  .208}{\textbf{+1.23}} & \textcolor[rgb]{ .329,  .51,  .208}{\textbf{+1.23}} & \textcolor[rgb]{ .329,  .51,  .208}{\textbf{+1.81}} \\
    \midrule
    \multirow{2}[2]{*}{Tokyo 24/7} & pretrained & 24.84 & 29.62 & 39.49 & 48.41 \\
          & \textbf{ICKD + MSE} & \textcolor[rgb]{ .329,  .51,  .208}{\textbf{+7.64}} & \textcolor[rgb]{ .329,  .51,  .208}{\textbf{+8.28}} & \textcolor[rgb]{ .329,  .51,  .208}{\textbf{+7.64}} & \textcolor[rgb]{ .329,  .51,  .208}{\textbf{+7.32}} \\
    \bottomrule
    \end{tabular}%
  \label{table:netvlad_lowres}%
  \vspace{-12pt}
\end{table}%

%% file: Parts/7-Conclusion.tex
\section{Conclusion}
Addressing the concern of reduced VPR accuracy under the realistic scenario of image quality reduction, our knowledge distillation methodology for extracting more discriminative descriptors from low quality images achieves significant performance gains under JPEG and video compression for various VPR methods and datasets. Furthermore, we have demonstrated our approach's potential to generalize to other kinds of low quality images such as low resolutions, reducing the literature gap on video-based VPR datasets in the process.

A suitable direction of future work is to justify distillation's effects on different kinds of VPR methods. While we have identified the general strengths of ICKD and MSE losses and noted the specific usability of the triplet loss, a better understanding of each VPR method's affinity to specific losses could reveal further insights on VPR methods themselves. Secondly, more VPR datasets and research under the context of video quantization could prove valuable for the future real-world applications of VPR.

\section*{Acknowledgment}

We would like to express our gratitude to Zezheng Li and Liyuan Geng for their valuable assistance with test data preprocessing, accelerating this work's progress.